\pdfoutput=1
\documentclass[11pt]{article}

\usepackage[final]{coling}

\usepackage{latexsym}

\usepackage[T2A]{fontenc}

\usepackage[utf8]{inputenc}

\usepackage{microtype}

\usepackage{inconsolata}

\usepackage[english, russian]{babel}

\usepackage{booktabs}
\usepackage{siunitx}
\usepackage{graphicx}
\usepackage{makecell}
\usepackage{xcolor}

\title{Low-resource Machine Translation for Code-switched Kazakh-Russian Language Pair}

\author{Maksim Borisov \\
  ITMO University \\ Saint-Petersburg, Russia \\
  \small{\texttt{borisovmaksim@niuitmo.ru }} \\\And
  Zhanibek Kozhirbayev  \\
Nazarbayev University \\ Astana, Kazakhstan \\ 
  \small{\texttt{zhanibek.kozhirbayev@nu.edu.kz}} \\\And
  Valentin Malykh  \\
  ITMO University \\ Saint-Petersburg, Russia \\
  \small{\texttt{valentin.malykh@phystech.edu}} \\}

\begin{document}

\renewcommand{\figurename}{Figure}
\renewcommand{\tablename}{Table}

\maketitle
\begin{abstract}
Machine translation for low resource language pairs is a challenging task. This task could become extremely difficult once a speaker uses code switching.
We propose a method to build a machine translation model for code-switched Kazakh-Russian language pair with no labeled data. Our method is basing on generation of synthetic data. Additionally, we present the first code-switching Kazakh-Russian parallel corpus and the evaluation results, which include a model achieving 16.48 BLEU almost reaching an existing commercial system and beating it by human evaluation.

\end{abstract}

\section{Introduction}

Code-switching presents a significant challenge in Natural Language Processing due to its unpredictability, variability, and the lack of available corpora, especially for low-resource languages. In this paper, we propose a method for training a machine translation system on the Kazakh-Russian language pair. There is no publicly available code-switched Russian-Kazakh parallel dataset, thus we present one in this work. This datasets contains only 600 parallel sentences, so it can be used only for evaluation and not for training. In our method we use several publicly available Russian-Kazakh datasets, but since these datasets do not address code-switching problem, we generate additional training data by translating relevant monolingual corpus and show the effectiveness of this approach. To augment the data we developed  a novel text transformation method based on SimAlign \cite{sabet2020simalign}. We used our generated data in thorough evaluation of the existing models. We fine-tune these models on the generated dataset resulting in 16.13 BLEU for the best baseline model, while Yandex commercial model shows 16.72 BLEU. These experimental results suggest that our method is able to improve the performance of machine translation systems on real code-switching data and jump start for those language pairs which has no code-switched data collected.

The following paper is structured as follows: section~\ref{sec:related} describes the works on code-switching done for other language pairs alongside with studies devoted to Russian-Kazakh language pair; section~\ref{sec:datasets} presents the description of the existing public datasets for mentioned language pair and the description of newly introduced dataset with code-switching phenomenon captured; section~\ref{sec:method} contains the details regarding our proposed augmentation method; section~\ref{sec:evaluation} describes the baselines, their training process and the achieved results, while section~\ref{sec:conclusion} concludes the paper.

The contribution of this work is three-fold: (i) We present the first Russian-Kazakh code-switching dataset; (ii) we present an evaluation of the existing models on this dataset; (iii) we propose a novel data augmentation for not code-switched datasets, allowed us to fine-tune the existing open models achieving almost on par performance with available commercial system.

\section{Related Work}
\label{sec:related}
% Zhanibek

Recent progress in NLP has spurred the development of technologies capable of handling code-switched data. Despite the initiation of Code-Switching research several years ago, progress within the research community has been sluggish. The primary challenges in addressing this issue arise from the insufficient availability of data (\citealp{winata-etal-2023-decades}). A limited number of languages, such as Spanish-English (\citealp{weller-etal-2022-end}; \citealp{xu-yvon-2021-traducir}), Hindi-English (\citealp{appicharla-etal-2021-iitp}; \citealp{jadhav-2022}), or Chinese-English (\citealp{li-etal-2012-mandarin}), dominate research and resources in CSW. Nevertheless, numerous countries and cultures that extensively use CSW remain underrepresented in NLP research. 

In the past few years, there has been a growing focus on various tasks utilizing code-switched data, encompassing Language Modeling, especially in the domains of Automatic Speech Recognition (ASR) (\citealp{pratapa-etal-2018-language}; \citealp{garg-etal-2018-code}; \citealp{gonen-goldberg-2019-language}; \citealp{winata-etal-2019-code}; \citealp{lee-li-2020-modeling}). Numerous pre-trained model strategies leverage multilingual language models like mBERT or XLM-R for processing Code-Switching data (\citealp{khanuja-etal-2020-gluecos}; \citealp{aguilar-solorio-2020-english}; \citealp{winata-etal-2021-multilingual}). These models are commonly fine-tuned with downstream tasks or CSW text to enhance adaptability across different languages. The primary focus of attention has been on Language Identification in code-switched data, largely attributed to the inaugural and subsequent Shared Tasks during EMNLP 2014 and 2016 (\citealp{solorio-etal-2014-overview}; \citealp{molina-etal-2016-overview}; \citealp{zubiaga2016tweetlid}). Other tasks in this domain are Named Entity Recognition (\citealp{aguilar-etal-2018-named}), Part-of-Speech tagging (\citealp{aguilar-etal-2020-lince}; \citealp{khanuja-etal-2020-gluecos}), and Sentiment Analysis (\citealp{patwa-etal-2020-semeval}). 

Despite the notable advancements in multilingual NMT, these models remain tailored for monolingual input and output at both ends of the translation. While there has been increased research attention on MT for code-switching in recent years, the efforts in this domain are still relatively limited. Historically, endeavors have predominantly focused on leveraging artificial code-switched data to enhance traditional translation systems. For example, \citet{huang-yates-2014-improving} employed CSW corpora to refine word alignment and statistical MT. \citet{dhar-etal-2018-enabling} worked towards developing a machine translation system for code-mixed content. They introduced a new parallel corpus for English-Hindi code-switched pairs. This corpus was utilized to improve existing machine translation systems by supplementing data in the parallel corpora. \citet{singh-2018} established a Hindi-English parallel corpus containing code-switching from Facebook data. The dataset also takes into consideration spelling variants in social media code-switching. In the context of English-Arabic, \citet{menacer-2019} curated a parallel corpus by extracting UN documents. This multilingual code-switching corpus encompasses not only English on the Arabic side but also French, Spanish, and other languages. \citet{dinu-etal-2019-training} enhanced term translation accuracy by replacing and concatenating source terminology constraints with corresponding translations. \citet{song-etal-2019-code} introduced "soft" constraint decoding by replacing phrases with pre-specified translations. They explored a data augmentation method, transforming code-switched training data by substituting source phrases with their corresponding target translations based on predefined translation lexicons.  Another avenue of research (\citealp{bulte-tezcan-2019-neural}; \citealp{xu-etal-2020-boosting}) explores methods to integrate a source sentence with similar translations extracted from translation memories. \citet{yang-etal-2020-csp} pre-trained translation models by predicting original source segments from generated CSW sentences, claiming superior results compared to other pretraining methods.

A common feature of natural interactions among bilingual speakers is the spontaneous and continuous switching between the Kazakh and Russian languages. It is worth noting that the field still faces challenges, particularly due to the scarcity of code-switched data and the colloquial characteristic of code-switching. To our knowledge, only a few research papers have been published on this matter. In the context of Kazakh-Russian code-switching, a study by \citet{ubskii2020impact} attempted to determine the benefit of bilingual training on matrix language (Kazakh) and embedded language (Russian) monolingual data \cite{myers1997duelling}, as opposed to training on code-switched data only. The study made use of two datasets: Kazakh speech with code-switching and Russian speech with no code-switching. The main objective of the experiments was to compare the performance of a model trained on code-switched speech with that of a model trained on full utterances in both languages. Experimental results suggested that bilingual training improves the model’s performance on matrix words, and greatly improves its performance on embedded words. Another study by \citet{zharkynbekova2022kazakh} discussed the ethnic bilingual practice in Kazakhstan. The focus was on code-switching or, in other term, code-mixing in the Kazakh-Russian and Russian-Kazakh bilingualism. The bi- and multilingualism is characteristic for Kazakhstan and is caused by multi-ethnicity of the republic. The study analyzed 300 contexts that show the Kazakh-Russian code-mixing in everyday and internet communication, and in modern Kazakh films reflecting the typical code-mixing practice.

\section{Datasets} 
\label{sec:datasets}
\subsection{Training Datasets}
Training datasets are consists of a dataset collected by Nazarbayev University and described in \cite{nu}, we refer to this dataset as NU below; a dataset collected by Al Farabi University and described in \cite{kaznu} (KazNU);
domain adaptation dataset, which is based on Russian tweet corpus described in \cite{RTC} (see Section~\ref{sec:method} for details). 
    These three datasets are the main sources of training data, in addition we use several smaller datasets. 
    %TODO: add description for these datasets.
    To acquire these datasets we used MTData tool described in \cite{gowda-etal-2021-many}.  We combine all the datasets in a single one and apply deduplication. We call this dataset ``all~data'' below.
    We provide the statistics for all the training datasets in Tab.~\ref{table:train}.

% добавить количество токенов в русском и казахском
\begin{table*}[tbh!]
    \centering 
    \caption{Train datasets statistics.}
    % \vspace{2mm} 
    \begin{tabular}{|l|ll|l|}
    \hline
        Dataset Name & \#Sentences & \#Ave. Tokens & Domain \\ \hline
        NU \cite{nu} & 895372 & 20.58 & Juridical docs \\ 
        KazNU \cite{kaznu} & 80627 & 20.74 & Off. press-releases \\ 
        Russian tweet corpus \cite{RTC} & 12752816 & 7.88 & Social media \\ 
        Statmt-news\_commentary-15-kaz-rus &  11735 & 19.43 & News \\ 
        Statmt-news\_commentary-14-kaz-rus & 9204 & 19.15 & News \\ 
        Statmt-news\_commentary-16-kaz-rus & 13224 & 19.42 & News \\ 
        Facebook-wikimatrix-1-kaz-rus & 165109 & 10.09 & Web docs\\ 
        OPUS-tatoeba-v2-kaz-rus & 2010 & 8.59  & General \\ 
        OPUS-wikimatrix-v1-kaz-rus & 32807 & 10.47 & Wikipedia \\ 
        OPUS-tatoeba-v20190709-kaz-rus & 2390  & 8.27& General  \\ 
        OPUS-tatoeba-v20210310-kaz-rus & 2401 & 8.26& General  \\ 
        OPUS-tatoeba-v20210722-kaz-rus & 2417 & 8.24 & General \\ 
        OPUS-multiccaligned-v1-kaz-rus & 1841440 & 4.94 & Web docs \\
        OPUS-xlent-v1.1-kaz-rus & 87167 & 2.05 & Software doc-n  \\ 
        OPUS-kde4-v2-kaz-rus & 68014 & 4.70 & Software doc-n \\ 
        OPUS-qed-v2.0a-kaz-rus & 5125 & 10.74&  Software doc-n \\ 
        OPUS-opensubtitles-v2016-kaz-rus & 1246 & 4.55 & Subtitles \\ 
        OPUS-ubuntu-v14.10-kaz-rus & 235 & 4.13 & Software doc-n  \\ 
        OPUS-wikimedia-v20210402-kaz-rus & 40714 & 16.41& Wikipedia  \\ 
        OPUS-tatoeba-v20200531-kaz-rus & 2400 & 8.26& General   \\ 
        OPUS-multiccaligned-v1.1-kaz-rus & 431952 & 12.04& Web docs \\ 
        OPUS-ted2020-v1-kaz-rus & 9484 & 12.05& Subtitles \\ 
        OPUS-opensubtitles-v2018-kaz-rus & 2223 & 4.21& Subtitles \\ 
        OPUS-news\_commentary-v14-kaz-rus & 9163 & 19.12& News \\ 
        OPUS-news\_commentary-v16-kaz-rus & 9163 & 19.03 & News\\ 
        OPUS-tatoeba-v20220303-kaz-rus & 2418 & 8.59& General  \\ 
        OPUS-xlent-v1-kaz-rus & 307929 & 2.05& Software doc-n  \\ 
        OPUS-gnome-v1-kaz-rus & 20550 & 3.07 & Software doc-n \\ 
        OPUS-tatoeba-v20201109-kaz-rus & 2401 & 8.26 & General  \\ \midrule
        all data (dedup.) & 20424090 &  & Mixed \\\hline
    \end{tabular}
    \label{table:train}
\end{table*}

\subsection{Evaluation Dataset}

\begin{table*}[tbh!]
    \centering 
    \caption{KRCS dataset statistics.}
    % \vspace{2mm} 
    \begin{tabular}{|l|l|}
    \hline
        %Dataset Name & \#Sentences & \#Ave. Tokens & Domain \\ \hline
        
        Number of sentences & 618 \\
        Ave. \# of tokens in an original Kazakh sentence & 11.95 \\ 
        Ave. \# of Russian tokens in an original sentence &  2.77 \\ 
        Ave. \# of tokens in a corrected Kazakh sentence & 12.27 \\ 
        Ave. \# of tokens in a Russian sentence & 13.64 \\ 
        
        \hline
    \end{tabular}
    \label{table:krcs}
\end{table*}

We use Kazakh-Russian Code-Switching dataset (KRCS) as our evaluation dataset. The KRCS dataset consists of 618 colloquial Kazakh sentences from social media which include some Russian phrases with corresponding ground truth translations to grammatically correct Kazakh and Russian labeled by annotators. %There are XX tokens in an included Kazakh sentence
We had two annotators, both of them were natively bilingual in Kazakh and Russian, both of them are working in academia. The annotation were done as part of their academic duties.

% Переделать пример
\begin{table*}
    \centering 

    % \vspace{2mm} 
    \begin{tabular}{|l|c|c|}
    \hline
        % Type & Sentence  \\ \hline
     Original &  \makecell{\textcolor{blue}{казахстанский} гендерлік теңдік саласындағы \\ 12 халықаралық құжаттарды бекітті .} & \makecell{\textcolor{blue}{фискал} көзқарастан \\ гөрі либералдандыру жақсы}
  \\  
        Corrected  & \makecell{Қазақстан гендерлік теңдік саласындағы \\ 12 халықаралық құжаттарды бекітті .} & \makecell{Фискалдық көзқарастан \\ гөрі либералдандыру жақсы}
  \\ 
         
  Russian &  \makecell{\textcolor{blue}{Казахстаном} ратифицировано  \textcolor{red}{12} \textcolor{teal}{международных} \\ документов в сфере  \textcolor{orange}{гендерного} равенства .} & \makecell{\textcolor{teal}{Лучше} \textcolor{red}{либерализация} ,\\ чем \textcolor{blue}{фискальный} \textcolor{orange}{подход}} \\
  \hline
    \end{tabular}
    \caption{Sample sentence triplet from KRCS dataset.}
    \label{table:krcs_sample}
\end{table*}

The descriptive statistics of the collected corpus is provided in Tab.~\ref{table:krcs}. In Tab.~\ref{table:krcs_sample} we provide a sample from KRCS dataset.

\section{Dataset Augmentation}
\label{sec:method}

\subsection{Code-Switching Emulation Method}
In the previous section we described the training datasets, nevertheless we need to state clearly that that datasets are not consider code-switching phenomenon and thus cannot be used effectively in our setup. Therefore we decided to make code-switching data artificially, using specific techniques for data augmentation.

First, we prepare the data. For it we follow the M2M100 recipe provided in \href{https://github.com/facebookresearch/fairseq/blob/main/examples/m2m_100/README.md}{fairseq repository} which is an official implementation of~\cite{fairseq}. Namely, we filter out sentences with more than 50\%  of punctuation, remove the duplicates, and discard sentences with more than 50\% of symbols that are not common for a given language.

Next, we take Kazakh processed sentences and augment them. We developed five different types of data augmentation to create code-switching training data. These are: 
\begin{itemize}
    \item cs-1: Replace a Kazakh word with a Russian one in normal form.
    \item cs-2: Replace a Kazakh word with a Russian one's stem with Kazakh ending, extracted from a Kazakh word by excluding stem from it.
    \item cs-3: Replace a Kazakh word with a Russian one in random form.
    \item cs-4: Replace a Kazakh word with a Russian word aligned using fastalign \cite{dyer2013simple}.
    \item cs-5: Replace a Kazakh word with a Russian word aligned using SimAlign \cite{sabet2020simalign}.
\end{itemize}

For cs-1, cs-2, and cs-3 we employ a publicly available Kazakh-Russian dictionary from work~\cite{rakhimova2020normalization}. % добавить цитату
 
For cs-4 and cs-5 Minimal Aligned Units are extracted following an approach described in~\cite{xu-yvon-2021-traducir}: the small billingual phrase pairs $(a, b)$ extracted from symmetrical alignment such that for every word in $a$ there exists a link to word in $b$ and vise versa.

For all augmentation methods, we replace 15\% of tokens in the Kazakh sentence at random\footnote{The exact percentage is inspired by Masked Language Modeling approach firstly introduced in \cite{devlin2018bert}}. Sentences with length of less than 7 tokens have one replacement following~\cite{anwar2023effect}. We provide samples for each augmentation type in Tab.~\ref{table:examples}. We also provide additional linguistic analysis and justification for each method in Appendix~\ref{sec:aug_ana}.

% The matrix language is fixed to be Kazakh and embedded language is Russian according to The Matrix Language Frame theory \cite{myers1997duelling}. % не понимаю, что это и зачем

\begin{table*}[tbh!]
    \centering 

    % \vspace{2mm} 
    \begin{tabular}{|l|c|c|}
    \hline
        % Type & Sentence  \\ \hline
        kk  & \makecell{Қазақстан гендерлік теңдік саласындағы \\ 12 халықаралық құжаттарды бекітті .} & \makecell{Фискалдық көзқарастан \\ гөрі либералдандыру жақсы}
  \\ 
         cs-1 &  \makecell{\textcolor{blue}{казахстанский} гендерлік теңдік саласындағы \\ 12 халықаралық құжаттарды бекітті .} & \makecell{\textcolor{blue}{фискал} көзқарастан \\ гөрі либералдандыру жақсы}
  \\  
   cs-2 & \makecell{\textcolor{blue}{казахстансктан} гендерлік теңдік саласындағы \\  12 халықаралық құжаттарды бекітті .} & \makecell{\textcolor{blue}{фискадық} көзқарастан \\ гөрі либералдандыру жақсы}
  \\ 
   cs-3 & \makecell{Қазақстан гендерлік теңдік саласындағы \\ 12 \textcolor{teal}{международной} құжаттарды бекітті .} & \makecell{Фискалдық көзқарастан \\ \textcolor{teal}{скорейших} либералдандыру жақсы}
  \\ 
cs-4 & \makecell{Қазақстан гендерлік теңдік саласындағы \\ \textcolor{red}{12} халықаралық құжаттарды бекітті .} & \makecell{Фискалдық көзқарастан \\ гөрі \textcolor{red}{либерализация} жақсы}
  \\ 
   cs-5 & \makecell{Қазақстан \textcolor{orange}{гендерного} теңдік саласындағы  \\ 12 халықаралық құжаттарды бекітті .} & \makecell{Фискалдық \textcolor{orange}{подход}  \\гөрі либералдандыру}
  \\ 
  ru &   \makecell{\textcolor{blue}{Казахстаном} ратифицировано  \textcolor{red}{12} \textcolor{teal}{международных} \\ документов в сфере  \textcolor{orange}{гендерного} равенства .} & \makecell{\textcolor{teal}{Лучше} \textcolor{red}{либерализация} ,\\ чем \textcolor{blue}{фискальный} \textcolor{orange}{подход}} \\
  \hline
    \end{tabular}
    \caption{Examples of code-switching augmentations}
    \label{table:examples}
\end{table*}

% Next the sentencepiece subword tokenization model is trained with shared vocabulary size of 32000 tokens \cite{kudo2021sentencepiece}. We discard sentence encodings with more then 250 tokens or less than 1 token and with max length / min length ratio over 1.5.

\subsection{Domain Adaptation}
\begin{figure*}[tbh!]
  \includegraphics[width=\linewidth]{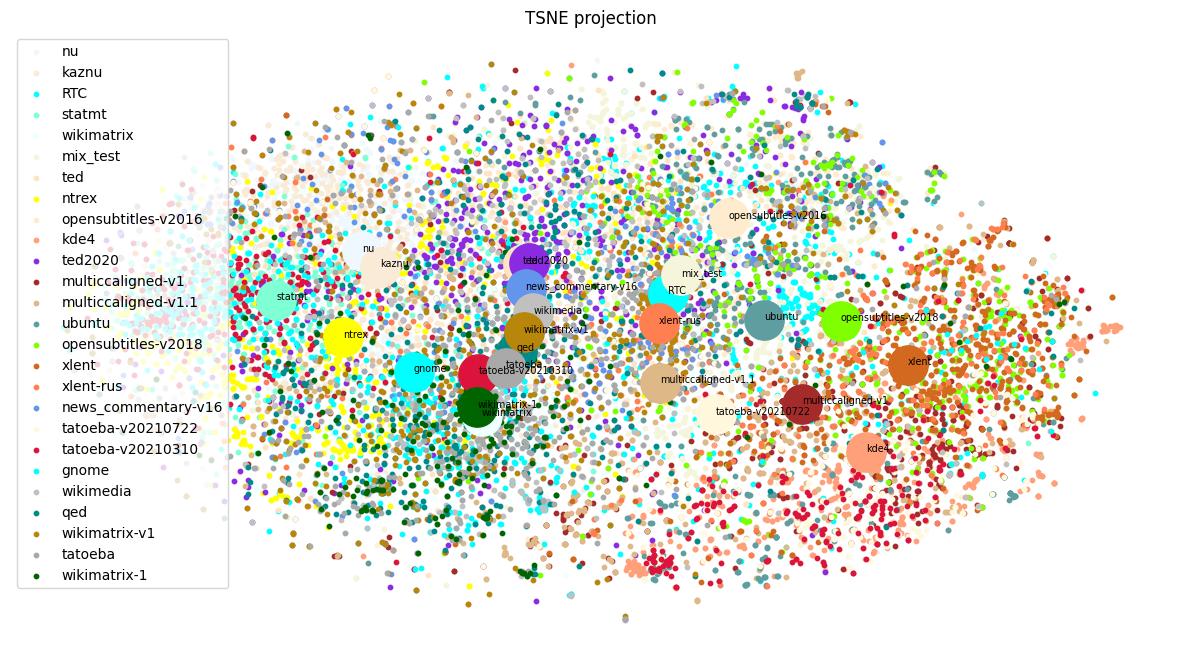}
   \caption{Sentence embedding visualization with dataset centroids.}
   \label{image:tsne}
\end{figure*}

As one can conclude from the previous section, there is a domain mismatch from the available training data and collected evaluation data. We provide a visualization of this mismatch in Fig.~\ref{image:tsne}.
It is a tSNE projection of LaBSE embeddings~\cite{labse} of the Kazakh sentences from the training datasets and Russian sentences from Russian tweet corpus. One can see that centroid of Russian Tweet Corpus is closer to the centroid of KRCS dataset than any other one of another dataset. This observation drove us to conclusion that we might need a domain adaptation. 

Since Russian tweet corpus is a monolingual Russian language dataset, we translated it to Kazakh using publicly available machine translation model nllb-200-distilled-600M from NLLB model family described in~\cite{nllb}. Our choice of the model was driven by the fact that it shows the best quality in standard Russian-Kazakh translation. In Tab.~\ref{table:train} for Russian tweet corpus we report number of Kazakh tokens for the generated translation. We call this corpus RTC throughout the paper. 

% KazNU test subset, KRCS code switching dataset, ntrex dataset \cite{ntrex} and tedtalks dataset \cite{opus}. See Table \ref{table:test}.

% \begin{figure}[t]
% \centering
% \caption{TSNE projection}
% \includegraphics[width=17cm]{latex/images/all_data_tsne.png}
% \label{image:tsne}
% \end{figure}

% \begin{table}
%     \centering 
%     \caption{Test datasets}
%     \begin{tabular}{|l|l|}
%     \hline
%      Dataset & Size \\ \hline
%       nu &  5000 \\ 
%       kaznu & 1000 \\ 
%       KRCS & 618 \\ 
%       Microsoft-ntrex-128-kaz-rus & 1997 \\ 
%       Neulab-tedtalks\_dev-1-rus-kaz &  \\
%       Neulab-tedtalks\_test-1-rus-kaz & 3987 \\
%       Neulab-tedtalks\_train-1-rus-kaz &  \\ \hline
%     \end{tabular}
%     \label{table:test}
% \end{table}

% \begin{table*}
%     \centering 
%     \caption{BLEU scores on KRCS dataset}
%     \label{table:mix_test}
% \begin{tabular}{|cc|}
%     \toprule
%       identity        & 7.55          \\
%     facebook/mbart-large-50-many-to-many-mmt & 4.62    \\
        
%     facebook/m2m100\_1.2B       &  5.37  \\

%     facebook/nllb-200-distilled-600M   &   12.26 \\
%     facebook/nllb-200-3.3B  & 15.23 \\
%     yandex translation   &   16.72 \\
%     google translation   &   24.14 \\
%     \bottomrule
% \end{tabular}

% \end{table*}

% Maxim

\section{Evaluation}
 \label{sec:evaluation}
 
 \subsection{Baselines}
 There are several baselines which are used in our experiments. We use \textbf{identity} baseline, which simply copying its input to the output. This baseline is obviously not trained.
 
 There are two trained from scratch baselines, namely, the first one is \textbf{transformer-600}, which is described below. The architecture of the model follows NLLB one, specifically the 600M parameters variant. It is implemented in fairseq framework~\cite{fairseq}. The model has 6 encoder layers and 6 decoder layers with hidden size of 512. Feed forward network hidden dimension is 4096, there are 8 attention heads for encoder and for decoder. Layer normalization before each encoder and decoder block is applied.  
%For regularization we apply dropout of 0.3, Attention dropout of 0.2 and ReLU dropout of 0.2 (which in a dropout probability after ReLU in FFN). %\cite{dropout} 
The embedding matrices for encoder input, decoder input and decoder output are all shared. The model was optimized using Adam~\cite{adam}.
% with betas of $(0.9, 0.98)$ and epsilon $1e^{-0.6}$. Scheduler is inverse square root with initial learning rate of $3e^{-0.5}$ and warmup of 2500 updates.  Max tokens per batch is 2048. Maximum number of updates is 500000. Criterion is label smoothed cross entropy  with smoothing factor of 0.2. \cite{smoothing}. 
The full list of hyperparameters can be found in Appendix~\ref{sec:app}.

The second trained from scratch baseline is a reproduced approach from~\cite{nu}. We call this baseline \textbf{transformer-NU}.

The next three baselines are using pre-trained machine translation models and fine-tune them on our training data. These baselines are \textbf{mBART}, a model family described in~\cite{bart}, we use specifically \texttt{mbart-large-50-many-to-many-mmt} variant; \textbf{M2M100}, a model family described in~\cite{m2m100}, specifically \texttt{facebook/m2m100\_1.2B}; and \textbf{NLLB-600}, a model family described in~\cite{nllb}, specifically \texttt{facebook/nllb-200-distilled-600M}.

The last fine-tuned baseline is \textbf{NLLB-3.3B} from the same model family as the previous one, but it is \texttt{facebook/nllb-200-3.3B} variant. We do not fully fine-tune this model, instead we use PiSSA~\cite{pissa}, a PEFT approach.

%It was trained on Kazakh - Russian language pair and
% experimental setup

\subsection{Metrics}
In our work we are using three standard metrics: BLEU score~\cite{bleu}, which is basically a token accuracy; ChrF++ score~\cite{chrf}, which is character level F-score; and COMET score~\cite{rei2020comet}, which is a Transformer-based model trained to compare translations. For the last metric we use specifically \texttt{Unbabel/wmt22-cometkiwi-da} model, described in~\cite{rei2022comet}.

\subsection{Augmentation Study}

\begin{table*}[tb!]
    \centering
    \begin{tabular}{|c|ccccccc|}
    \toprule
        Data & NU & CS-1 & CS-2 & CS-3 & CS-4 & CS-5 & KRCS \\ \hline
        all data  & 36.03 & 33.10 & 33.41 & 33.00 & 29.64 & 35.16 & 12.25 \\ 
        all data $+$ cs-1 & 35.07 & \textbf{34.34} & 33.60 & \textbf{33.67} & 30.51 & 34.25 & 10.20 \\ 
        all data $+$ cs-2 & 35.54 & 33.78 & \textbf{35.17} & 33.49 & 30.03 & 34.52 & 11.65 \\ 
        all data $+$ cs-3 & 34.24 & 33.37 & 32.94 & 33.25 & 29.53 & 33.42 & 10.22 \\ 
            all data $+$ cs-4 & 35.58 & 32.87 & 33.10 & 32.74 & \textbf{33.69} & 37.03 & 11.38 \\ 
        all data $+$ cs-5 & \textbf{36.83} & 33.68 & 34.18 & 33.63 & 32.96 & \textbf{39.05} & \textbf{12.49} \\ 
        \bottomrule
    \end{tabular}
    \caption{The BLEU scores for transformer-600 model on differently augmented datasets.}
    \label{table:csw_test}
\end{table*}

\begin{table*}[bt!]
    \centering
    \scalebox{0.8}{

    \begin{tabular}{|c|ccccccc|}
    \toprule
        Data & NU & CS-1 & CS-2 & CS-3 & CS-4 & CS-5 & KRCS \\ \hline
        all data  & 61.28 / 0.82& 59.58 / 0.76 & 59.44 / 0.76 & 58.96 / 0.75 &  56.73 / 0.69 & 61.78 / 0.78 & \textbf{37.10} / 0.52  \\ 
        all data $+$ cs-1 & 60.64 / 0.81 & \textbf{60.13 / 0.77} & 59.67 / 0.77 & 59.51 / 0.76 & 56.95 / 0.69 & 60.47 / 0.77 & 34.52 / 0.51 \\ 
        all data $+$ cs-2 & 61.09 / 0.82  & 59.67 / 0.77 & \textbf{60.85 / 0.78} & 59.53 / 0.76  & 56.76 / 0.69 & 61.12 / 0.77 & 36.02 / 0.52 \\ 
        all data $+$ cs-3 & 60.33 / 0.81 & 59.62 / 0.77  & 59.50 / 0.77 & \textbf{59.59 / 0.76} & 56.18 / 0.69 & 59.82 / 0.77 & 33.72 / 0.50 \\ 
            all data $+$ cs-4& 59.81 / 0.81 & 58.16 / 0.74 & 58.33 / 0.74 & 58.16 / 0.74 & \textbf{59.15 / 0.69} & 62.56 / 0.77 & 34.81 / 0.51 \\ 
        all data $+$ cs-5 & \textbf{61.63 / 0.82} & 59.22 / 0.75 & 59.68 / 0.75 & 59.20 / 0.74 & 58.81 / 0.69 &  \textbf{64.27 / 0.79} & 36.44 / \textbf{0.54} \\ 
        \bottomrule
    \end{tabular}
    }
    \caption{The ChrF++ and COMET scores for transformer-600 model on differently augmented datasets.}
    \label{table:csw_test2}
\end{table*}

In this section we provide a comparison for the models trained on different augmentation types. We train our transformer-600 model on cs-1, cs-2, cs-3, cs-4 and cs-5 augmented datasets. The detailed description of these augmentations can be found in section~\ref{sec:method}. We evaluate the trained models on testing subset of NU dataset, and its augmented  versions. A version of NU test set augmented with  by cs-1 is called CS-1, the other types are called in the same manner. More importantly we evaluate the models on KRCS dataset. The results are presented in Tables~\ref{table:csw_test}\&\ref{table:csw_test2}.

Interesting, that the only augmentation type which helps to improve the baseline results is cs-5. All other types are leading to decrease in quality. For all the types, except cs-3, the evaluation on corresponding augmented testset is the best. For cs-3 the best result is achieved by a model trained on CS-1, this result is not surprising since the cs-3 augmentation is just a random choice between cs-1 and cs-2 augmentations. % cs-3 надо переделввать, т.к. это должна быть случайная форма русского слова
 Another interesting point is that cs-5 augmentation allowed a model to achieve the best performance on the original testset. We hypothesize, that this augmentation produces the closest data distribution to the spoken Kazakh language, thus effectively extending the trainset.

%Next, we fine-tune existing models on all datasets augmented by cs-5 strategy. We consider transformer baseline \cite{nu}, mbart-large-50-many-to-many-mmt, facebook/m2m100\_1.2B, facebook/nllb-200-distilled-600M and facebook/nllb-200-3.3B. 

%Furthermore, we provide results on KRCS dataset from yandex translation and google translation commercial API's in table \ref{table:mix_test} 

% Maxim
 % 
\begin{table*}[tbh!]
    \centering 
\begin{tabular}{|c|cc|}
    \toprule
    Model & w/o training & trained \\\midrule
      identity       & 7.55 / 25.10 / \textbf{0.56} & N/A        \\
    transformer-NU & 7.87 / 31.99 /  0.50 & 11.31 / 35.35 / 0.53\\
    transformer-600 & N/A  &   12.49 / 36.44 / 0.54\\
    mBART & 4.62 / 17.83 / \textbf{0.56} &  12.08 / 34.31 /  0.53 \\
        
    M2M100       &  5.37 / 21.59 / 0.42 &  12.50 / 36.44 / 0.53 \\

    NLLB-600   &   12.26 / 36.67 / 0.53 &  12.95 / 36.44 / 0.54 \\
    NLLB-3.3B  & \textbf{15.23} /\textbf{ 39.68} / \textbf{0.56} & \textbf{16.48} / \textbf{42.27} / \textbf{0.56 }\\\midrule
    \multicolumn{3}{|c|}{Commercial APIs} \\\midrule
    Yandex MT$^2$   &   22.24 / 47.13 / 0.67 & N/A \\
    Google MT   &   24.14 / 47.84 / 0.64 & N/A \\
    \bottomrule
\end{tabular}
 \caption{The comparison of baseline models in BLEU / ChrF++ / COMET on KRCS dataset.}
    \label{table:mix_test}

\end{table*}

\section{Results}
For this evaluation we use all the baselines with cs-5 augmentation, since it is the best in our setup as it shown in previous section. In addition, we provide results for two commercial machine translation systems, namely Yandex MT and Google MT. The results are provided in Tab.~\ref{table:mix_test} As one can see, the best results are achieved by NLLB-3.3B model. This is not surprising, once it is the biggest model in comparison. What is interesting in this setup is that our approach allows to achieve good results with all the trained models, and the best trained model achieves a score close to Yandex MT system\footnote{We provide current scores for Yandex MT system at the day of submission, namely 15th of June. When the work has been started the score for Yandex MT was \textbf{16.72} BLEU.}. Another point worth mentioning that COMET scores are close for identity baseline, mBART model, and NLLB-3.3B model.
%We have found that adding augmented RTC corpus to training data resulted in $+0.64$ BLEU.

%As for code-switching augmentations, training on cs-1 augmentations gave the best score on synthetic testing dataset created by cs-1 augmentations. The same goes for cs-2, cs-4 and cs-5 augmentations. For the cs-3 augmentation, the best score is obtained on model trained with cs-1 augmentation. This result is not surprising since the cs-3 augmentation is just a random choice between cs-1 and cs-2 augmentations.

%The best result on KRCS is obtained by novel cs-5 augmentation, giving $+0.24$ BLEU. Also, the cs-5 augmentations have given the best score on nu test dataset. It is an interesting effect. One possible explanation could be that by providing accurate alignment, the mbert model transfers some knowledge to our model. 

%Our proposed method applied to several known translation models via fine-tuning, improves their scores. We improve baseline by $4.3$ BLEU, mbart-large-50-many-to-many-mmt  by $7.46$ BLEU, m2m100\_1.2B  by $7.13$ BLEU and nllb-200-distilled-600M by $0.69$ BLEU. 

%We report the results for  facebook/nllb-200-3.3B model in Table \ref{table:mix_test} but do not fine-tune it due to resource limitations.

% надо добавить оценку переводов человеком, а также оценку аугментации
\subsection{Human Evaluation}
We have done human evaluation for the best baseline model and commercial APIs. We asked our assessors to use Likert scale and averaged their scores for 100 random sentences from KRCS. The results are provided in Tab.~\ref{table:human}. As can be seen, the results are a bit unexpected. Despite the automatic metrics scoring the Yandex MT system higher than NLLB-3.3B model, human evaluation showed the opposite. Also, it is worth noting that even the best commercial system is pretty far from ground truth translation in this domain.

We also evaluated the naturalness of augmentation in Kazakh. We chose 100 random sentences with cs-5 augmentation and asked our assessors again to use Likert scale. The achieved result is 2.62, which could be considered acceptably acceptable.

\begin{table}[h!]
    \centering
    \begin{tabular}{|c|cc|}
    \toprule
    & Mean & \small{Std.} \\
    \midrule
        Ground Truth & 4.75 & \small{0.68} \\ 
        NLLB-3.3B & 3.09 & \small{1.13} \\ 
        Yandex MT  & 2.80 & \small{1.17} \\ 
        Google MT & 3.49 & \small{1.14} \\ 
        \bottomrule
    \end{tabular}
     \caption{The human evaluation results.}
    \label{table:human}
\end{table}

\subsection{Domain Adaptation Evaluation}
We decided to evaluate the importance of domain adaptation corpus which is extend our training dataset. We trained our  transformer baseline model three setups, namely: all the training data, including RTC, all the training data, \textit{ex}cluding RTC, and RTC only. The experiments show that domain adaptation is indeed important, but the single domain adaptation data is not enough to achieve high performance in code switching task. These results are in Tab.~\ref{table:rtc}.

\begin{table}[htb!]
    \centering
    \begin{tabular}{|c|c|}
    \toprule
   
        Data & KRCS \\   \hline
        all data & 12.25 / 37.10 / 0.52 \\ 
        all data  w/o RTC & 11.64 / 35.58 / 0.49 \\ 
        RTC only & 10.86 / 34.76 / 0.52\\ 
        \bottomrule
    \end{tabular}
     \caption{The results of training on different datasets.}
    \label{table:rtc}
\end{table}

\section{Conlusion}
\label{sec:conclusion}
%In this study, we propose two methods of improving code-switching  machine translation for low-resource setting. Firstly, we make use of relevant monolingual corpus, creating additional training data. 
%Secondly, we introduce a new code-switching augmentation.
%Our experiments confirm the effectiveness of proposed methods. 

In conclusion, the proposed method demonstrates a viable approach to tackling machine translation challenges for low-resource, code-switched language pairs, specifically Kazakh-Russian. By utilizing synthetic data generation, the method circumvents the need for labeled training data, which is typically scarce for such language pairs. 

Furthermore, the introduction of the first code-switching Kazakh-Russian parallel corpus represents a significant contribution to the field, providing a valuable resource for future research and development. The empirical results, with the best developed model achieving a BLEU score of 16.48, indicate that the system’s performance is nearly on par with, and in some aspects, surpasses that of existing commercial translation systems, as evidenced by superior human evaluation outcomes. This highlights the effectiveness and potential of the proposed approach for improving machine translation in similar low-resource, code-switched contexts.

\section{Limitations}
Synthetic Data Dependence: The approach relies heavily on the generation of synthetic data, which may not perfectly capture the nuances and complexities of natural code-switching in Kazakh-Russian speech.

Evaluation Scope: While achieving a BLEU score of 16.48 is promising, the evaluation is limited to specific criteria and doesn't necessarily account for all aspects of translation quality, such as fluency and contextual accuracy.

Corpus Size and Diversity: The newly presented code-switching Kazakh-Russian parallel corpus may still be limited in size and diversity, potentially impacting the generalizability of the model to broader linguistic contexts or different dialects.

Commercial System Comparison: The performance comparison to an existing commercial system is based on certain benchmarks and human evaluations, which might not cover all practical use cases and scenarios where the commercial system might excel.

Scalability and Adaptability: The method's scalability to other low-resource, code-switched language pairs is not addressed, raising questions about its broader applicability and adaptability to different linguistic environments.

Long-term Sustainability: There is no discussion on the long-term sustainability and maintenance of the synthetic data generation process and how it might evolve with changes in the language pair dynamics or increased data availability.

By acknowledging these limitations, future research can focus on addressing these gaps to further enhance the robustness and applicability of machine translation models for code-switched languages.

\section{Ethical Considerations}
1. Privacy and Data Security: The development of the code-switching Kazakh-Russian parallel corpus and synthetic data generation involves processing potentially sensitive linguistic data. Ensuring that data is anonymized and securely stored is critical to protect the privacy of individuals.

2. Bias in Synthetic Data: The reliance on synthetic data may introduce biases that do not reflect real-world language use. It's essential to consider the potential for these biases to affect the fairness and accuracy of the translation system, particularly for marginalized or underrepresented communities.

3. Cultural Sensitivity: Code-switching often involves culturally significant language use. It’s important to ensure that the translation model respects and properly handles cultural contexts, avoiding misinterpretations that could lead to misunderstandings or misrepresentations.

4. Transparency and Accountability: The creation and use of the synthetic data and parallel corpus should be transparent, with clear documentation and methodologies made available to the research community. This transparency enables accountability and facilitates peer review and reproducibility.

5. Impact on Language Use: Machine translation systems influence how languages are used and perceived. Introducing a translation system for a low-resource, code-switched language pair might inadvertently affect the natural evolution of the languages involved or the prevalence of code-switching in the community.

6. Potential Misuse: The developed translation model could be misused, for instance, to create deceptive or misleading content. Ensuring that appropriate safeguards and ethical guidelines are in place when deploying such technologies is crucial.

7. Equity in Accessibility: Ensuring equitable access to this translation technology is essential, particularly for the Kazakh and Russian-speaking communities that might benefit most from it. Efforts must be made to avoid digital divides and ensure that the system is accessible to diverse user groups.

By addressing these ethical considerations, developers and researchers can work towards creating a more responsible and beneficial machine translation system.

\bibliography{main}

\appendix
\section{Hyperparameters}
\label{sec:app}

The hyperparameters of the Transformer-600 model are presented in Tab.~\ref{table:hyps}.
\begin{table}[tbh!]
    \centering 
\begin{tabular}{|cc|}
    \toprule
      Number of layers        & 6            \\
    Hidden size       &  512  \\
    FFN hidden dimension & 4096    \\
    Attention heads   &   8 \\
    LN before blocks  & True \\
    Max Tokens   &   2048 \\
    Criterion   &   label smoothed CE \\
    Label smoothing   &   0.2 \\
    Optimizer   &   adam \\
    Adam epsilon   &   1e-06 \\
    Adam betas   &   (0.9, 0.98) \\
    Lr scheduler   &   inverse sqrt \\
    Lr   &   3e-05 \\
    Warmup updates   &   2500 \\
    Dropout   &   0.3 \\
    ReLU dropout   &   0.2 \\
    Attention dropout   &   0.2 \\
    Share all embeddings   &   True \\
    Max update  &   500000 \\

    \bottomrule
\end{tabular}
    \caption{Model Hyperparameters. LN stands for Layer Normalization. CE stands for Cross-Entropy.}
    \label{table:hyps}

\end{table}

\section{Augmentation Analysis}
\label{sec:aug_ana}

\paragraph{cs-1:} Replace a Kazakh word with a Russian one in normal form
Linguistic 

\textit{Soundness}: This approach is straightforward and resembles natural code-switching seen in everyday speech, where speakers often insert words from another language in their base form, especially nouns and technical terms.

\textit{Examples}: In Kazakh media and daily conversations, you might hear sentences like “Мен жаңа ручка сатып алдым” (“I bought a new pen”), where “ручка” is a Russian-origin word used in its normal form.

\textit{Usage Contexts}: Such patterns are common in informal speech, especially when referring to modern or technical terms for which there might be no direct equivalent in Kazakh.

\paragraph{cs-2:} Replace a Kazakh word with a Russian word’s stem with Kazakh ending

\textit{Linguistic Soundness}: This is somewhat less natural, as it involves morphologically adapting Russian stems with Kazakh endings, which does not always fit the natural phonological or morphological rules of Kazakh. However, speakers often perform such blending to maintain grammatical consistency within a sentence.

\textit{Examples}: This is occasionally seen in youth slang or creative language use in social media where Kazakh speakers playfully adapt Russian words. For instance, “жазать” (from Russian “писать” but adapted to sound more Kazakh) might appear in informal texts, though not formally accepted.

\textit{Usage Contexts}: This type of adaptation is mostly informal, often perceived as a playful or creative linguistic exercise rather than standard usage.

\paragraph{cs-3:} Replace a Kazakh word with a Russian one in random form

\textit{Linguistic Soundness:} This approach might lack naturalness as it disregards context, grammar, and sentence flow. The randomness can introduce syntactic or morphological anomalies.

\textit{Examples:} You might hear mismatched forms in spontaneous bilingual speech, particularly among less proficient speakers who switch languages mid-sentence without full grammatical integration. For example, “Мен пошел домой” ("I went home" mixing Kazakh and Russian), where the Russian verb form is not conjugated correctly according to Kazakh syntax.

\textit{Usage Contexts: }Common in highly informal settings, such as among bilingual children or learners who are not fully competent in both languages.

\paragraph{cs-4:} Replace a Kazakh word with a Russian word aligned using FastAlign
\textit{Linguistic Soundness:} Using statistical alignments like FastAlign generally improves the naturalness of word replacements because it considers contextual word pairs frequently appearing together in parallel corpora.

\textit{Examples:} News broadcasts or bilingual podcasts often use consistent patterns of switching, aligning with how FastAlign might map Kazakh-Russian sentence structures. For example, “Менің ойымша, это не совсем правильно” ("I think this is not quite right") frequently occurs.

\textit{Usage Contexts:} Seen in media content where consistent patterns in code-switching reflect translation or repeated bilingual interactions.

\paragraph{cs-5:} Replace a Kazakh word with a Russian word aligned using SimAlign

\textit{Linguistic Soundness:} SimAlign uses contextual embeddings, making this approach more linguistically sound as it considers sentence-level semantics for alignment. This tends to produce contextually appropriate and grammatically fitting replacements.

\textit{Examples:} In digital content, such as YouTube videos or podcasts with bilingual speakers, there are instances like “Бұл өте интересно тақырып” (“This is a very interesting topic”), where alignment mirrors natural bilingual communication.

\textit{Usage Contexts:} Common in both formal and informal settings, particularly where speakers frequently shift between languages without disrupting the overall meaning.

\end{document}